\documentclass[sigconf,nonacm]{acmart}
\usepackage{graphicx}
\usepackage{subcaption}
\usepackage{multirow}
\usepackage{adjustbox}
\usepackage{tablefootnote} 
\usepackage{relsize}
\usepackage{float}
\AtBeginDocument{%
  \providecommand\BibTeX{{%
    \normalfont B\kern-0.5em{\scshape i\kern-0.25em b}\kern-0.8em\TeX}}}

\acmBooktitle{} 
\acmPrice{}
\acmISBN{}

\begin{document}

\title[Interpretable AD in Cellular Networks by Learning Concepts in VAEs]{Interpretable Anomaly Detection in Cellular Networks by Learning Concepts in Variational Autoencoders}

\author{Amandeep Singh\textsuperscript{1}}
\email{amandeep.b.singh@erocsson.com}
\affiliation{%
  \institution{Ericsson Services GmbH}
  \streetaddress{Prinzenallee 21}
  \city{Düsseldorf}
  \country{Germany}
  \postcode{40549}
}

\author{Michael Weber\textsuperscript{2}}
\email{michael.weber@ericsson.com}
\affiliation{%
  \institution{Ericsson Services GmbH}
  \streetaddress{Prinzenallee 21}
  \city{Düsseldorf}
  \country{Germany}
  \postcode{40549}
}

\author{Markus Lange-Hegermann\textsuperscript{3}}
\email{markus.lange-hegermann@th-owl.de}
\affiliation{%
  \institution{inIT – Institut für industrielle Informationstechnik,
  Ostwestfalen-Lippe University of Applied Sciences}
  \streetaddress{Campusalle 12, 32657}
  \city{Lemgo}
  \country{Germany}}

\renewcommand{\shortauthors}{Amandeep, et al.}

\begin{abstract}
This paper addresses the challenges of detecting anomalies in cellular networks in an interpretable way and proposes a new approach using variational autoencoders (VAEs) that learn interpretable representations of the latent space for each Key Performance Indicator (KPI) in the dataset. This enables the detection of anomalies based on reconstruction loss and Z-scores. We ensure the interpretability of the anomalies via additional information centroids $(c)$ using the K-means algorithm to enhance representation learning. We evaluate the performance of the model by analyzing patterns in the latent dimension for specific KPIs and thereby demonstrate the interpretability and anomalies. The proposed framework offers a faster and autonomous solution for detecting anomalies in cellular networks and showcases the potential of deep learning-based algorithms in handling big data. 
\end{abstract}





\maketitle

\section{Introduction}
The use of mobile technology has exploded over the last few decades, with the growth of telecommunications ranging from simple half-duplex voice communication to smartphones \cite{solyman2022evolution}. This expanding industry not only represents a large business sector but also has a significant impact on society. The recent COVID-19 pandemic has highlighted the critical role that digital infrastructure plays in society, with mobile networks serving as an essential means of staying connected and working from home. According to GSMA's Mobile Economy Report \cite{intelligence2023mobile}, by 2030, the number of unique mobile subscribers is projected to reach 6.3 billion, showing an increase from 5.4 billion in 2022. This indicates an additional 0.9 billion individuals subscribing to mobile phone services. The report also predicts a mobile penetration rate of 73\% by the end of 2030, compared to 68\% in 2022. Notably, the report anticipates that 5G technology alone will account for 54\% of all mobile connections by 2030, surpassing half of the total mobile subscriptions. This massive expansion of mobile technology usage generates an enormous amount of statistical data on a minute, hourly, and daily basis. 

Statistical data plays a crucial role in enabling network operators to initiate troubleshooting efforts promptly when necessary. This data encompasses both normal and abnormal behavior observed in cellular networks. Domain experts rely on statistical data to monitor network performance and identify underperforming or anomalous areas for further improvement. However, manually monitoring this vast amount of statistical data to detect anomalous behavior is both costly and time-consuming for network operators. 
To expedite and automate the tedious process of anomaly detection in mobile networks, several methods have been developed. These methods include evaluating intrusion detection systems for mobile devices using machine learning classifiers \cite{damopoulos2012evaluation}, predicting time-series anomalies and mobile malware \cite{dandekar2023approach, sharma2022approach}, exploring anomaly detection techniques specifically designed for mobile networks \cite{9124843}. Other methods include finding feature distribution outliers \cite{stoecklin2006anomaly}, diagnosing device-specific anomalies by applying entropy-based analysis in cellular networks \cite{schiavone2014diagnosing}, utilizing a distribution-based approach for anomaly detection in 3G mobile traffic \cite{d2009distribution}, and employing big data analytics for anomaly detection in cellular networks \cite{li2019anomaly}. Furthermore, a survey paper \cite{chalapathy2019deep} highlights the effectiveness of deep learning-based algorithms in detecting anomalies on large-scale data, which was for example successfully applied to cellular mobile networks \cite{al2019lstm}.

Regardless of different anomaly detection methods, there are challenges in accurately detecting anomalies in cellular networks. It is very difficult to define a common boundary for any observation that deviates from the normal behavior of each network element. Another challenge is identifying the exact Key Performance Indicators (KPIs) that cause anomalies in the cellular network dataset. For instance, anomalies in the KPIs can manifest as a swift increase in call drop rate and a sudden decrease in downlink and uplink (DL/UL) throughput for a specific network element in a cellular network. 

To address these challenges, we employed the variational autoencoder (VAE) \cite{kingma2013auto,doersch2016tutorial} that learns a probability distribution that represents the values of the KPIs in a given dataset. We demonstrate that VAE can learn to represent {\itshape concepts} built for an individual KPI based on domain supervision in the given dataset. These concepts both improve the accuracy of the VAE when detecting anomalies and also describes which KPIs differ when an anomaly is detected. The latter point is particularly important to interpret and thereby fix the anomalies.

Each {\itshape concept} is a multivariate Gaussian corresponding to a set of datapoints. We choose these sets to contain all data that is similar in a specific KPI. The formation of conceptual labels for given datasets is explained in second section. In the domain of images, concepts bundle a set of images having a similar color, size, shape etc., as discussed in \cite{shaikh2022conceptual}. It is worth noting that we could easily construct these concepts in a conditional variational autoencoder \cite{sohn2015learning, walker2016uncertain}, incorporating them as additional information (one-hot encoding) in the input of the encoder and decoder. However, in the vanilla and conditional variational autoencoders, the explicit representation of the concept is lacking, hence probabilistic methods of pin-pointing anomalies is impossible. In our model, the incorporation of a concept in the prior of a particular dimension of the latent space turns out to be the key difference.

The interpretable latent dimensions facilitate gaining insights into the detected anomalous points. This enables the utilization of unsupervised methods for predicting anomalies based on loss measures, while also leveraging representation learning to assess the identified anomalies. Additionally, having a concise representation of dimensions proves valuable for cellular network dataset due to the significance of mobility within the network. The performance of a network element is greatly influenced by its neighboring elements, and a clear representation of the latent dimensions provides additional information about significant or insignificant changes in the surrounding area of an anomalous point.

The rest of the paper is organized as follows: In Section \ref{sec:approach}, we present our approach in detail. Section \ref{sec:data} is dedicated to data basis and experimental setup. Section \ref{sec:results} provides results of the proposed solution. We conclude in Section \ref{sec:conclusion}.

\section{APPROACH}
\label{sec:approach}
A VAE models the probability distribution of any given dataset using neural networks. Hence, it has the ability to generate new data and provides the likelihood of any new datapoint to belong to the probability distribution modelled by the VAE. The neural networks are consisting of encoder and decode, both of which are trained together and optimized via backpropagation. This framework includes a latent space, also known as the bottleneck, which compresses high-dimensional data into a low dimensional space and learns an inexplicable latent space representation. However, this intricate representation poses a problem when there is a need to interpret or represent the latent representation.

Previous studies have indicated that a stronger constraint on the latent bottleneck enhances the model's ability to learn efficient data representations \cite{higgins2017beta}. Another study suggests that incorporating anomaly labels at the bottleneck improves the reliability of anomaly detection in VAE \cite{hammerbacher2021including}. Furthermore, the conceptual VAE approach \cite{shaikh2022conceptual} has demonstrated that integrating additional information in the form of concepts can yield explicit representations in the latent dimensions. 

Building upon these findings, we first employ the k-means algorithm \cite{macqueen1967classification} to derive conceptual labels denoted as $(c)$ by forming clusters based on the mean values of each KPI for an individual network element in our time series dataset. This clustering technique is also applied to extract patterns from high-dimensional radio network performance datasets by \cite{wang2021extracting}. The centroids obtained from our dataset by the k-means clusters are incorporated as additional information at the bottleneck of the VAE during training, facilitating the learning of discernible latent dimensions. Our goal is to group network elements with low variance and an optimal number of clusters. To achieve this, we compare different clustering algorithms, including hierarchical clustering \cite{bindra2017detailed} and the k-means algorithm. Our observations reveal that the k-means clustering algorithm exhibits low homogeneity within the clusters, making it particularly suitable for our chosen dataset.

In a state-of-the-art VAE, the prior distribution plays a crucial role in controlling the latent distribution. This control allows the VAE to capture the complex patterns of the data in lower-dimensional latent space and learn meaningful representations. While latent dimensions themselves hard to interpret \cite{locatello2019challenging}, they encode valuable internal representations of observed data. However, our approach aims for interpretable representations of latent dimensions that can quantify the origin and magnitude of the captured anomalies.

\begin{table*}[ht]
  \centering
  \caption{Training, Validation and Testing dataset}
  \label{tab:dataset}
  \begin{tabular}{cccccc}
    \toprule
    \textbf{Dataset} & \textbf{KPIs} & \textbf{Centroids} & \textbf{Training} & \textbf{Validation} & \textbf{Testing} \\
    \hline
    Radio\_Call\_Drop & 5 & 5 & 11156584 & 587189 & 1985451 \\
    \bottomrule
  \end{tabular}
\end{table*}

To achieve this, we introduce a modification in our implementation by computing the Kullback-Leibler (KL) divergence between the encoder distribution and a specific condition $c$, the concept, instead of using a standard Gaussian. This results in a modified KL prior, represented as $p(z|c)$, rather than the standard Gaussian prior $p(z)$. We incorporate the centroids into the mean of the prior for a particular latent dimension, while keeping the standard deviation constant for all latent dimensions. This approach aims to fit the encoder distributions corresponding to $c$, allowing the model to learn distinct representations for specific concepts. Specifically, the mean of the KL prior varies based on the concept $c$ for a specific latent dimension at the bottleneck, while the remaining latent dimensions are encoded with a standard normal distribution $N(0,I).$ These modifications in the VAE model enhance interpretability by capturing and quantifying anomalies in terms of their magnitude, while simultaneously leveraging the power of Gaussian distributions in the latent space. Overall, our approach follows the three hypotheses in \cite{niggemann2015diagnosis}: (i) we follow a data-driven approach, (ii) that captures the timing, and (iii) we do not concentrate on root cause analysis but identify the failing component.

Furthermore, we utilize the learned Gaussian distributions in the latent dimensions to interpret predicted anomaly based on a high loss. To evaluate the anomaly caused by a specific KPI, we convert the mean value of latent dimensions into the Z-scores. Alternatively, one could also convert the Z-scores into p-value and evaluate the predicted anomaly based on them. However, in our setup, we calculate the Z-scores using the learned mean and standard deviation from the latent dimensions of the training dataset for each KPI.

The Z-score defines the upper and lower bounds for each KPI, which can be determined either based on the training dataset (normal dataset) or by using a hard-cut line according to domain knowledge. In our case, we set the hard-cut line at +15 for the Z-score of each KPI when evaluating the anomalies in section \ref{sec:Anomaly_detection}. This Z-score allows us to identify which specific KPI caused the anomaly and magnitude of the anomaly predicted by the model.

\section{DATA BASIS AND EXPERIMENTAL SETUP}
\label{sec:data}
We collected a time series dataset in cellular networks consisting of 5 KPIs: Call Drop Rate, Total Drops, eNodeB Drops, MME Drops, and Total Call Attempts. Monitoring these KPIs is essential for identifying and addressing instances of failures or anomalies promptly. They provide crucial information about the network status and the performance of cellular sites, serving as fundamental indicators to distinguish between anomalous and normal behavior in the network.

For our study, we selected a homogeneous area covering approximately 100,000 km\textsuperscript{2}, which is served by over 9,000 base stations with around 94,000 sector IDs. Each sector generates one data point per day, resulting in a dataset based on a 150-day observation period for training, validation, and testing on a daily basis as shown in Table \ref{tab:dataset}. It is important to note that due to intermittent periods of being off-air or new installations/activations during this time, some base stations do not provide data for the full 150 days.

Additionally, we incorporated 500 arbitrary centroids into the dataset. These centroids were obtained using the k-means algorithm based on domain knowledge, low variance, and an optimal number of clusters. Each centroid value falls within the range of -1 to 1. The dataset is a time-series and sequential nature, making it well-suited for recurrent neural network analysis.

The architecture of encoder comprises three LSTM layers with sigmoid and hyperbolic tangent activation functions, along with one linear layer directly before the bottleneck. Similarly, the decoder consists of three LSTM layers followed by one linear layer and sigmoid function deployed at the output, which helps to make the learning process more stable. Additionally, the exponential function is used to ensure positive log variance at both the output and bottleneck layers. To capture the variability of data, 25 latent dimensions are encoded with a standard normal distribution $N(0,I)$, along with 5 latent dimensions for the concepts derived from the KPIs. To enhance the bottleneck representation quality \cite{higgins2017beta}, a weighting term of 10 is assigned for the reconstruction loss. We initialize the weights using orthogonal initialization, ensuring that the recurrent weights have orthogonal matrices, while the biases are initialized with zeros. The input sequence length for all LSTM layers is set to 100. To optimize the neural networks' attributes, we utilize the Adam gradient-based optimizer, which optimizes stochastic objective functions using estimates of lower-order moments. The learning rate for the optimizer is set to 0.001.

\begin{figure}[ht]
  \centering
  \includegraphics[width=\linewidth]{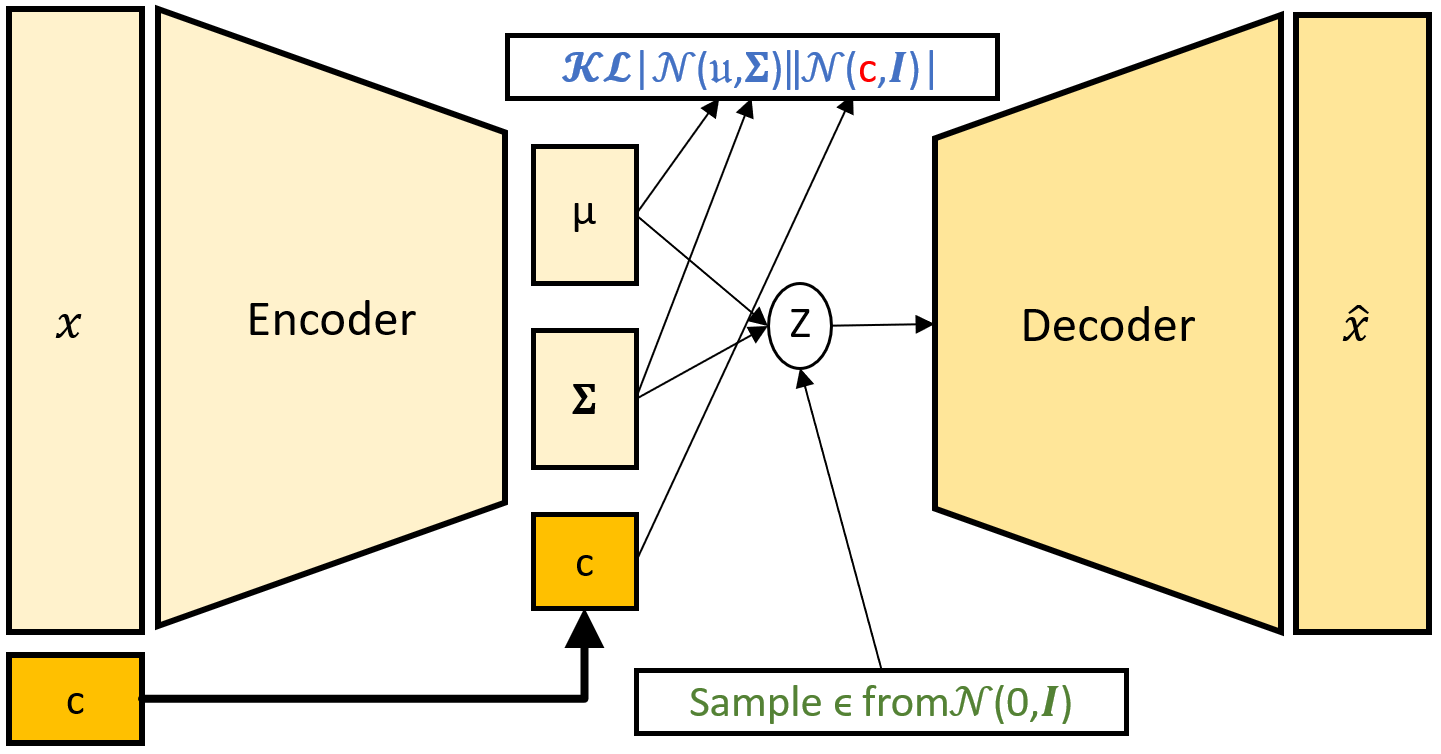}
  \caption{Variational Autoencoder incorporated with centroids $(c)$ from domain supervision}
  \label{fig:latent_space_fig}
\end{figure}

Fig. \ref{fig:latent_space_fig} illustrates the incorporation of the concept $c$ into the variational probability model. The input of the encoder $(x)$ and decoder $(Z)$ is independent of $c$, and the training proceeds similarly to the vanilla VAE. However, the main difference lies in the KL loss calculation, where the $c$ replaces the normal prior in the vanilla VAE. The model learns an individual Gaussian distribution for each concept $c$.
\begin{figure*}[ht]
  \centering
  \begin{subfigure}[b]{0.49\linewidth}
    \includegraphics[width=\linewidth]{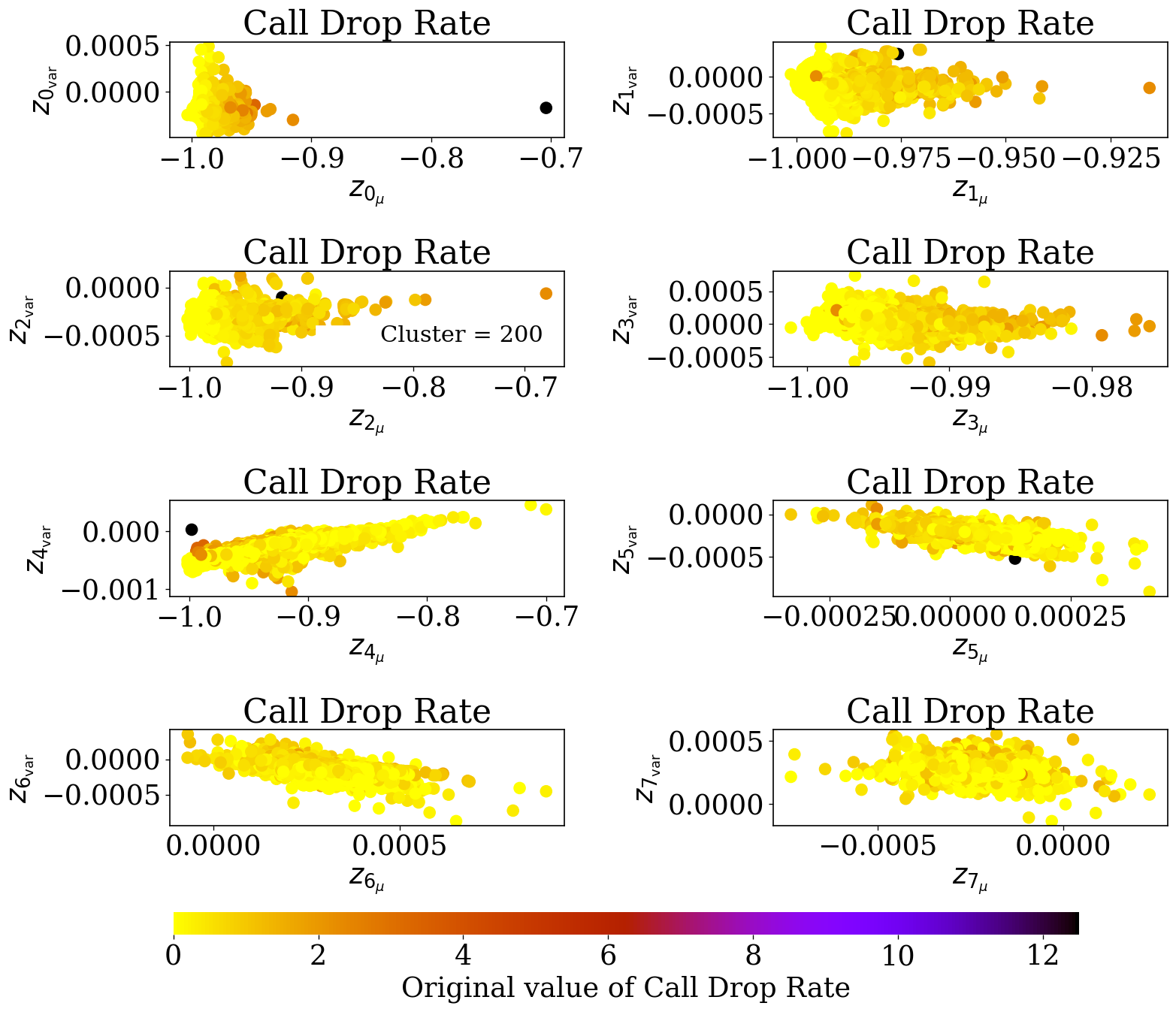}
  \end{subfigure}
  \hfill
  \begin{subfigure}[b]{0.49\linewidth}
    \includegraphics[width=\linewidth]{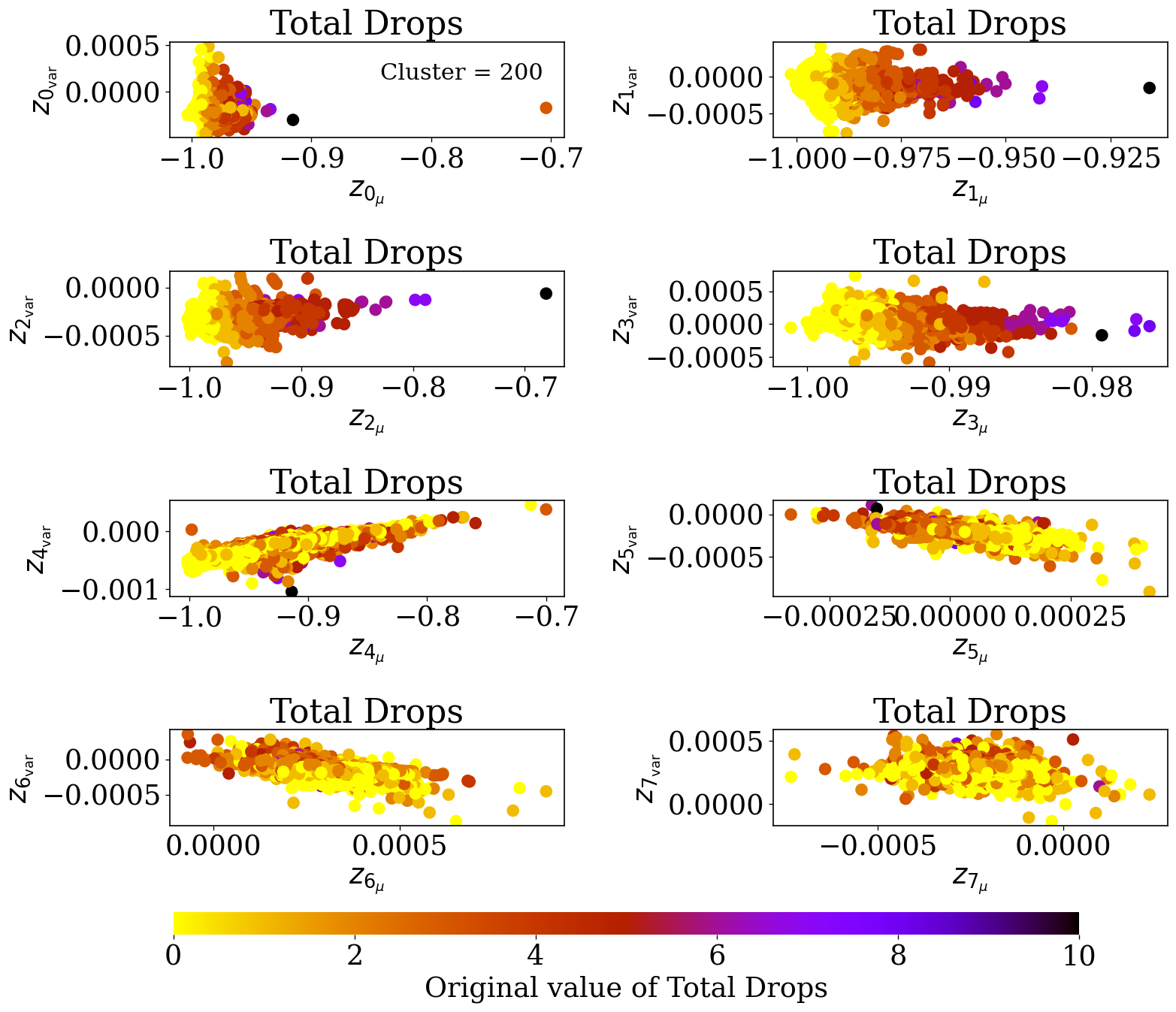}
  \end{subfigure}
  \caption{Latent representation of leaned means (x-axis) and log-variances (y-axis) predicted by the encoder for Call Drop Rate $(p(z0|c0))$ and Total Drops $(p(z1|c1))$ in training dataset for cluster 200}
  \label{fig:call_total}
\end{figure*}

We use the reconstruction probability method \cite{an2015variational}, which computes the log-likelihood defined by the mean and log variance drawn at the output of decoder. Our loss function is the usual $ELBO$ (evidence lower bound), which allows to jointly train the decoder and encoder networks. After training, the $ELBO$ is a good approximation to the likelihood. To better approximate the likelihood, we sample independently ten times in the latent space and then take the mean of the samples. The KL loss works as a regularizer and encodes the samples taken from a distribution centered around zero, but in our implementation, the encoded samples correspond to conceptual priors $p(z|c)$.

\begin{equation}
\text{Loss}_{KL} = \text{KL}(q(z|\mathbf{x})|| p(z|\mathbf{c})) 
\end{equation}

\begin{equation}
\text{Loss}_{\text{log-likelihood}} = \mathbb{E}_{z \sim q(z|\mathbf{x})} [\log p(\mathbf{x}|\mathbf{z})] 
\end{equation}

\begin{equation}
\text{$ELBO$} = \text{Loss}_{KL} - \text{Loss}_{\text{log-likelihood}}
\end{equation}

\section{RESULTS}
\label{sec:results}
This chapter shows the experimental results. First, we evaluate the latent representation of latent dimensions predicted by the encoder’s means and log-variances. After that, we evaluated the anomaly not only based on loss function but also learned Gaussian distributions from the latent representation.

\subsection{Representation Learning}
\begin{figure}[ht]
  \centering
  \includegraphics[width=\linewidth]{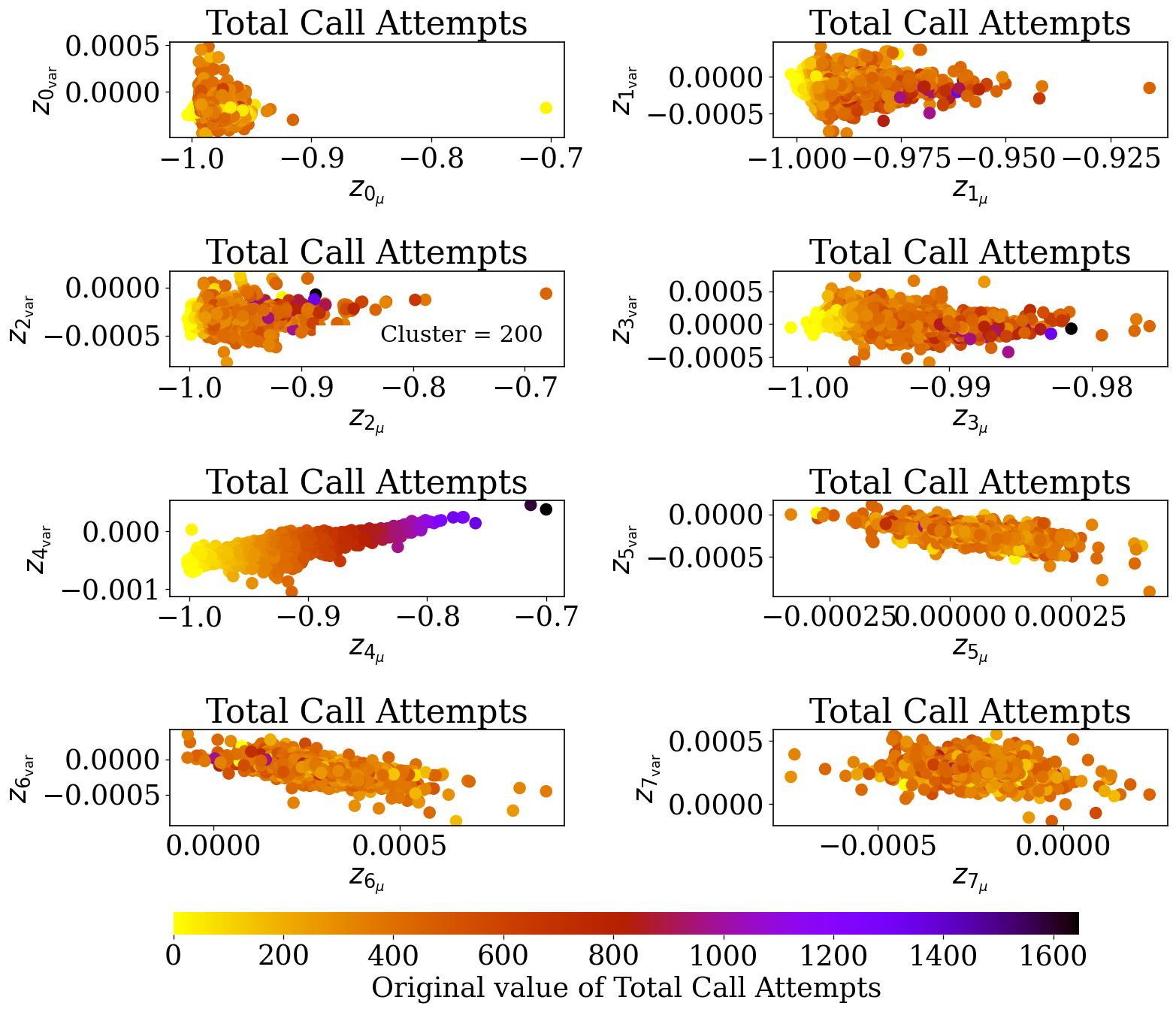}
  \caption{Latent representation of leaned means (x-axis) and log-variances (y-axis) predicted by the encoder for Total Call Attempts $(p(z4|c4))$ in training dataset for cluster 200}
  \label{fig:attempt}
\end{figure}

Fig. \ref{fig:call_total}, \ref{fig:attempt}, \ref{fig:enod_fig} and \ref{fig:mme_fig} illustrate the latent representation for eight dimensions. Among them, dimensions $z0$ to $z4$ encode five KPIs with conceptual priors, while the remaining three dimensions out of 25 latent dimensions that use a standard normal distribution. The visualization excludes the remaining latent dimensions since they are encoded with a standard normal distribution. To evaluate the results, we randomly selected cluster 200 out of a total of 500 clusters.

As anticipated, the left plot of Fig. \ref{fig:call_total} depicts a clear representation of the Call Drop Rate in the $z0$ dimension. Cluster 200 predominantly consists of data points close to zero, with only a few exceeding a value of two. The model effectively distinguishes higher values in the latent dimension, facilitating categorization of anomaly size. The right plot exhibits a distinct representation of Total Drops in the $z1$ dimension. Interestingly, a similar clear representation of Total Drops is also observable in the $z2$ dimension, as the eNodeB Drops KPI serves as a substitute for Total Drops. However, the other dimensions do not exhibit discernible representations.

\begin{table*}[ht]
  \centering
  \caption{Top 10 anomalies detected by the model for cluster 200 in the testing dataset}
  \label{tab:anomaly_result}
  \adjustbox{max width=\textwidth}{
    \begin{tabular}{|c|c|c|c|c|c|c|c|c|c|c|c|c|c|}
      \hline
      \multirow{2}{*}{\centering Cluster} & \multicolumn{5}{c|}{\centering Original values for all five KPIs} & \multicolumn{3}{c|}{\centering All Losses} & \multicolumn{5}{c|}{\centering Z-score for latent dimensions\footnote{* Values of latent dimensions show the high Z-score for a particular KPI in the table}} \\
      \cline{2-14}
      & Call Drop Rate & Total Drops & eNodeB Drops & MME Drops & Total Call Attempts & Loss & Log Likelihood & KL Loss & $z_{0_{\mathlarger{\mu}}}$ & $z_{1_{\mathlarger{\mu}}}$ & $z_{2_{\mathlarger{\mu}}}$ & $z_{3_{\mathlarger{\mu}}}$ & $z_{4_{\mathlarger{\mu}}}$ \\
      \cline{1-14}
      200 & 28.82 & 100 & 1 & 99 & 446 & 5.18 & -4.73 & 0.45 & \textbf{114.5*} & \textbf{111.8*} & 8.2 & \textbf{320.5*} & -0.8 \\
      200 & 24.94 & 96 & 1 & 95 & 480 & 5.12 & -4.71 & 0.41 & \textbf{118.1*} & \textbf{99.9*} & 2.6 & \textbf{295.9*} & 0.5 \\
      200 & 1.96 & 29 & 13 & 16 & 1497 & 4.72 & -4.60 & 0.12 & 9.3 & \textbf{35.2*} & \textbf{27.8*} & \textbf{50.0*} & 9.1 \\
      200 & 4.14 & 13 & 13 & 0 & 314 & 4.69 & -4.60 & 0.09 & \textbf{24.1*} & \textbf{21.1*} & \textbf{26.4*} & 10.1 & -0.4 \\
      200 & 4.6 & 30 & 2 & 28 & 680 & 4.68 & -4.60 & 0.07 & \textbf{27.6*} & \textbf{51.4*} & 8.8 & \textbf{138.2*} & 3.5 \\
      200 & 2.94 & 12 & 12 & 0 & 408 & 4.67 & -4.60 & 0.07 & \textbf{17.9*} & \textbf{19.7*} & \textbf{24.3*} & 10.5 & 0.7 \\
      200 & 13.33 & 8 & 0 & 8 & 68 & 4.67 & -4.60 & 0.06 & \textbf{65.3*} & 11.9 & 3.2 & \textbf{28.5*} & \textbf{-3.3} \\
      200 & 3.61 & 15 & 12 & 3 & 419 & 4.67 & -4.60 & 0.07 & \textbf{16.0*} & \textbf{23.1*} & \textbf{22.9*} & \textbf{23.2*} & 2.0 \\
      200 & 1.66 & 22 & 5 & 17 & 1344 & 4.66 & -4.60 & 0.05 & 8.3 & \textbf{28.4*} & \textbf{15.4*} & \textbf{54.9*} & 8.4 \\
      200 & 1.31 & 18 & 6 & 12 & 1388 & 4.65 & -4.60 & 0.06 & 6.0 & \textbf{22.0*} & \textbf{16.8*} & \textbf{32.7*} & 9.2 \\
      \hline
    \end{tabular}
  }
\end{table*}

\begin{figure*}[ht]
  \centering
  \begin{subfigure}[b]{0.49\linewidth}
    \includegraphics[width=\linewidth]{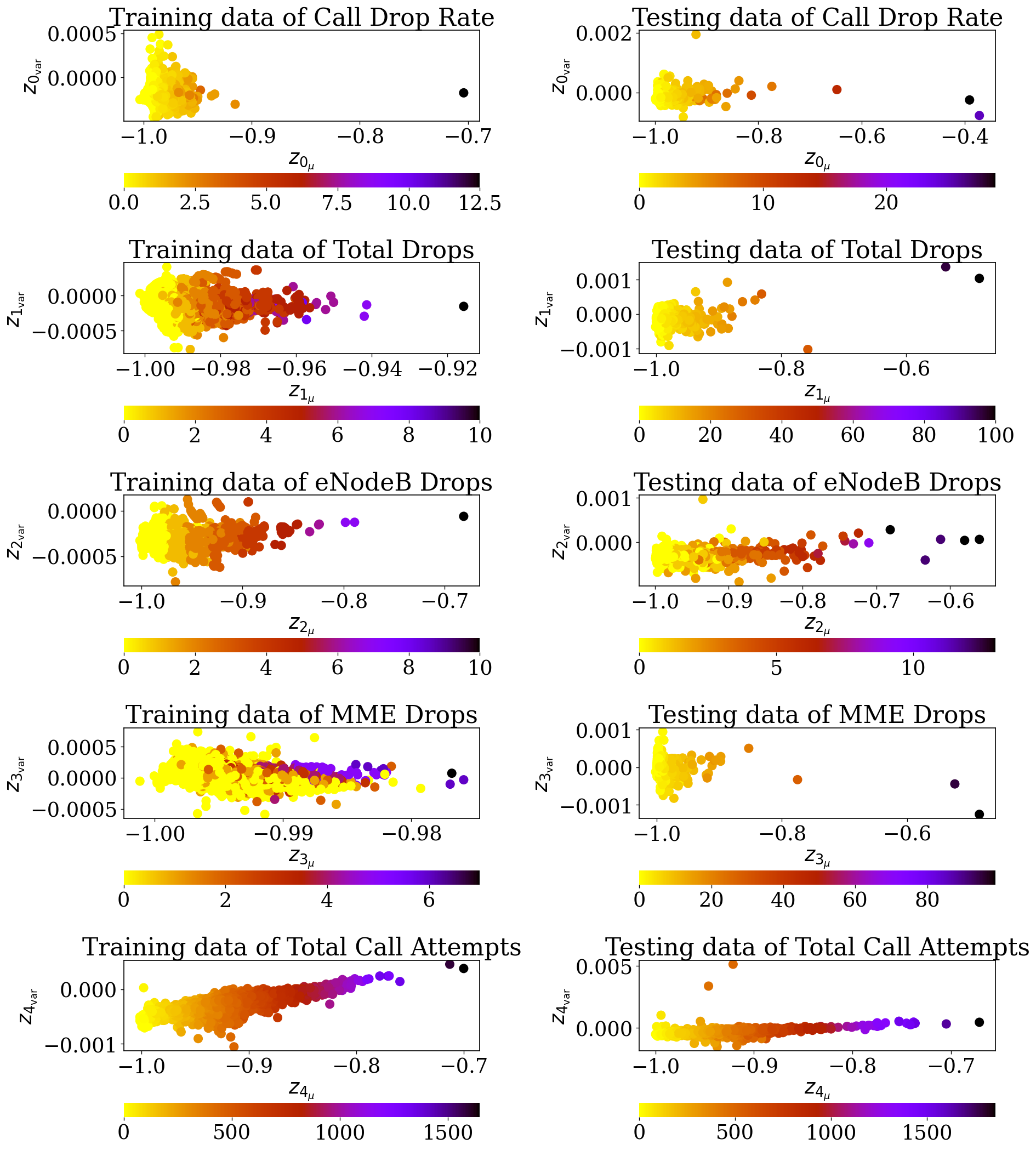}
  \end{subfigure}
  \hfill
  \begin{subfigure}[b]{0.49\linewidth}
    \includegraphics[width=\linewidth]{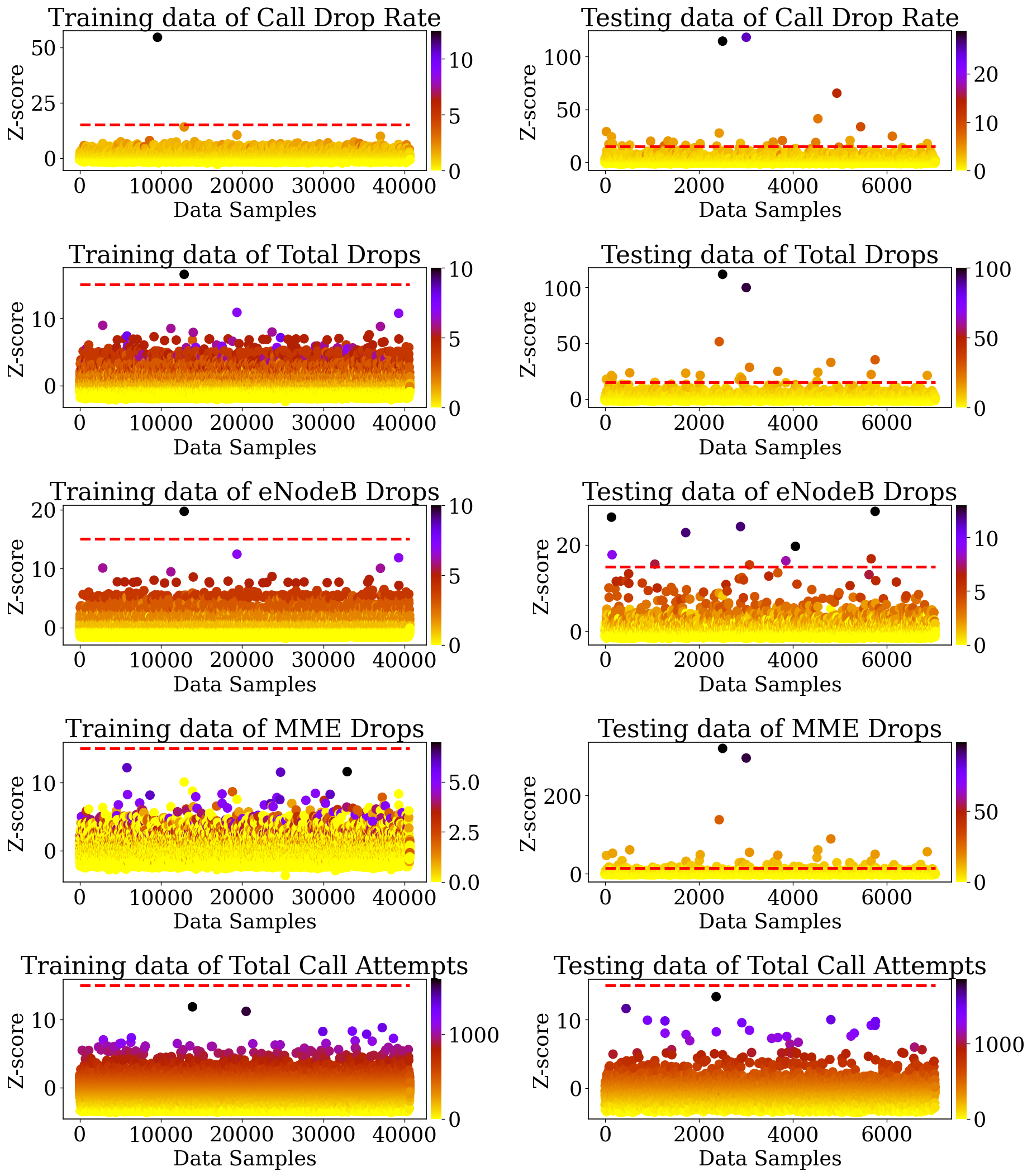}
  \end{subfigure}
  \caption{Left-plot shows latent representation of training and testing dataset, right-plot reveals anomaly detection on latent dimensions by Z-score for cluster 200}
  \label{fig:anomaly_concept}
\end{figure*}

Moving to the Fig. \ref{fig:enod_fig} in the appendix, we can observe the mapping of eNodeB Drops onto the eight latent dimensions. As expected, a clear representation is obtained in the $z2$ dimension. Additionally, the $z1$ dimension reveals a well-ordered representation of eNodeB Drops, albeit in the opposite direction due to the high correlation between the two KPIs. Fig. \ref{fig:mme_fig} in the appendix  presents the representation of MME Drops in the $z3$ dimension. While the learned dimension for this KPI is not as neat as for the other KPIs, higher values are still distinctly separated. Notably, the x-axis has a slight mean value ranging between -1 and -0.98, mapping the KPI values from 0 to 7. The variance is low for this KPI, and more samples in this specific cluster are close to zero compared to other KPIs. Although the model did not learn an elegant representation for this KPI, it excels in representing Total Call Attempts, as shown in Fig. \ref{fig:attempt}. The data points on the $z4$ dimension exhibits a linear increase. Importantly, a high variance is observed for this KPI, with data points distributed across all values. Therefore, the model effectively captures a linear representation when the KPI exhibits high variance.

\subsection{Anomaly Detection Evaluation}
\label{sec:Anomaly_detection}
The model learns the significant representation for each KPI corresponding to their conceptual priors. Hence, we could now use the learned Gaussian distributions obtained from the output of the encoder to interpret and determine the size of the predicted loss-based anomaly. The maximum value of some KPIs is higher in the test dataset (which comprises big anomalies) compared to the training dataset. Therefore, we use the min-max clipping from the training dataset to normalize the test data so that model predicts loss-based anomalies accurately. First, we investigate the KPIs representation for the training and testing dataset. In Fig. \ref{fig:anomaly_concept} we can see the change on the x-axis scale for all five dimensions between the training and testing datasets. The x-axis value is increased for the test dataset due to an increase in original values, which are potentially anomalies. Notably, the test dataset comprises the big anomalies and therefore $z3$ dimension shows a neat representation of the MME Drops; however, this is not the case for the training dataset. This is strong evidence that the model can penalize big anomalies more accurately. Besides this, the representation of $z0$, $z1$, $z2$ and $z4$ dimensions in the training dataset reveals that one or two separate data point on the x-axis is likely to be an anomaly. This visualization of data samples helps to clean up the training data set when the model needs to be trained only on the normal dataset.

After examining the representations of latent dimensions, we evaluate the entire test dataset and select only the top 10 predicted anomalies (loss > 4.65) based on the loss functions for cluster 200. Table \ref{tab:anomaly_result} illustrates that the model predicts anomaly points with a high loss value for cluster 200. Since each KPI is encoded corresponding to the concept $c$ and has its own univariate Gaussian distribution, we compute the Z-score for all five KPIs from $z0$ to $z4$ dimensions, respectively.

The Z-score allows us to determine which specific KPI is responsible for the predicted anomaly by the model. For example, the first row in Table \ref{tab:anomaly_result} shows that the Z-score is high only for the Call Drop Rate, Total Drops, and MME Drops KPIs, while the eNodeB Drops and Total Call Attempts KPI values are close to the learned mean value for their respective latent dimensions. This performance measure helps identify the correlation between KPI anomalies and isolates a particular anomalous KPI, facilitating the identification and resolution of the underlying problem. The Z-score information also aids in quickly identifying the performance of neighboring network elements. For instance, if the Z-score increases or decreases for a specific KPI in the surrounding area of the anomaly point, it enables domain experts to correlate the cause of the anomaly and identify the root causes behind the degradation of KPI.

Representation learning provides transparency for the predicted anomalous points. Therefore, one can use unsupervised methods to predict loss-based anomalies and utilize representation learning to evaluate the predicted anomalies. The right plot in Fig. \ref{fig:anomaly_concept} demonstrates the approach for detecting anomalies in the latent dimensions. The model may not predict the loss-based anomaly in case of a slight change in a specific KPI. However, the neat representation of the latent dimension can capture even a small change in that KPI. The red line at +15 is used to identify the anomaly points for each latent dimension. Data sample indexes are mapped on the x-axis, and Z-scores are on the y-axis for all five KPIs. The color bar scale indicates the original values for each KPI in dataset. In the test dataset, the color bar scale values and Z-scores are high for all five KPIs due to the presence of anomalous data samples. Additionally, the representation learning of the latent dimension can provide an individual Gaussian distribution for each network id, as cluster 200 consists of approximately 40,000 data samples belonging to around 300 network ids in the training dataset. Having an individual Gaussian distribution for each network id can assist in accurately capturing slight deviations.

\section{CONCLUSION}
\label{sec:conclusion}
This paper introduces an optimization framework for enhancing anomaly detection methods in cellular networks by incorporating additional information as priors into the VAE framework. The learned representations of each KPI allow us to assess and interpret predicted anomalies based on their reconstruction loss. The Gaussian distribution learned from the latent space provides clarity in identifying anomalous points and allows us to establish upper and lower bounds using the Z-score for specific conceptual labels. These bounds ultimately determine whether the anomaly is attributed to a high or low value in the KPI. Therefore, the representation learning of KPIs presents a compelling idea both theoretically and practically, offering a novel approach to anomaly identification. The experimental results validate the effectiveness of the VAE model in capturing the latent space and evaluating anomalies using the learned Gaussian distributions.

\bibliographystyle{ACM-Reference-Format}
\bibliography{myref}


\begin{thebibliography}{25}


\ifx \showCODEN    \undefined \def \showCODEN     #1{\unskip}     \fi
\ifx \showDOI      \undefined \def \showDOI       #1{#1}\fi
\ifx \showISBNx    \undefined \def \showISBNx     #1{\unskip}     \fi
\ifx \showISBNxiii \undefined \def \showISBNxiii  #1{\unskip}     \fi
\ifx \showISSN     \undefined \def \showISSN      #1{\unskip}     \fi
\ifx \showLCCN     \undefined \def \showLCCN      #1{\unskip}     \fi
\ifx \shownote     \undefined \def \shownote      #1{#1}          \fi
\ifx \showarticletitle \undefined \def \showarticletitle #1{#1}   \fi
\ifx \showURL      \undefined \def \showURL       {\relax}        \fi
\providecommand\bibfield[2]{#2}
\providecommand\bibinfo[2]{#2}
\providecommand\natexlab[1]{#1}
\providecommand\showeprint[2][]{arXiv:#2}

\bibitem[Al~Mamun and Beyaz(2019)]%
        {al2019lstm}
\bibfield{author}{\bibinfo{person}{SM~Abdullah Al~Mamun} {and}
  \bibinfo{person}{Mehmet Beyaz}.} \bibinfo{year}{2019}\natexlab{}.
\newblock \showarticletitle{LSTM recurrent neural network (RNN) for anomaly
  detection in cellular mobile networks}. In \bibinfo{booktitle}{\emph{Machine
  Learning for Networking: First International Conference, MLN 2018, Paris,
  France, November 27--29, 2018, Revised Selected Papers 1}}. Springer,
  \bibinfo{pages}{222--237}.
\newblock


\bibitem[An and Cho(2015)]%
        {an2015variational}
\bibfield{author}{\bibinfo{person}{Jinwon An} {and} \bibinfo{person}{Sungzoon
  Cho}.} \bibinfo{year}{2015}\natexlab{}.
\newblock \showarticletitle{Variational autoencoder based anomaly detection
  using reconstruction probability}.
\newblock \bibinfo{journal}{\emph{Special lecture on IE}} \bibinfo{volume}{2},
  \bibinfo{number}{1} (\bibinfo{year}{2015}), \bibinfo{pages}{1--18}.
\newblock


\bibitem[Bindra and Mishra(2017)]%
        {bindra2017detailed}
\bibfield{author}{\bibinfo{person}{Kamalpreet Bindra} {and}
  \bibinfo{person}{Anuranjan Mishra}.} \bibinfo{year}{2017}\natexlab{}.
\newblock \showarticletitle{A detailed study of clustering algorithms}. In
  \bibinfo{booktitle}{\emph{2017 6th international conference on reliability,
  infocom technologies and optimization (trends and future
  directions)(ICRITO)}}. IEEE, \bibinfo{pages}{371--376}.
\newblock


\bibitem[Chalapathy and Chawla(2019)]%
        {chalapathy2019deep}
\bibfield{author}{\bibinfo{person}{Raghavendra Chalapathy} {and}
  \bibinfo{person}{Sanjay Chawla}.} \bibinfo{year}{2019}\natexlab{}.
\newblock \showarticletitle{Deep learning for anomaly detection: A survey}.
\newblock \bibinfo{journal}{\emph{arXiv preprint arXiv:1901.03407}}
  (\bibinfo{year}{2019}).
\newblock


\bibitem[D'Alconzo et~al\mbox{.}(2009)]%
        {d2009distribution}
\bibfield{author}{\bibinfo{person}{Alessandro D'Alconzo},
  \bibinfo{person}{Angelo Coluccia}, \bibinfo{person}{Fabio Ricciato}, {and}
  \bibinfo{person}{Peter Romirer-Maierhofer}.} \bibinfo{year}{2009}\natexlab{}.
\newblock \showarticletitle{A distribution-based approach to anomaly detection
  and application to 3G mobile traffic}. In \bibinfo{booktitle}{\emph{GLOBECOM
  2009-2009 IEEE Global Telecommunications Conference}}. IEEE,
  \bibinfo{pages}{1--8}.
\newblock


\bibitem[Damopoulos et~al\mbox{.}(2012)]%
        {damopoulos2012evaluation}
\bibfield{author}{\bibinfo{person}{Dimitrios Damopoulos},
  \bibinfo{person}{Sofia~A Menesidou}, \bibinfo{person}{Georgios Kambourakis},
  \bibinfo{person}{Maria Papadaki}, \bibinfo{person}{Nathan Clarke}, {and}
  \bibinfo{person}{Stefanos Gritzalis}.} \bibinfo{year}{2012}\natexlab{}.
\newblock \showarticletitle{Evaluation of anomaly-based IDS for mobile devices
  using machine learning classifiers}.
\newblock \bibinfo{journal}{\emph{Security and Communication Networks}}
  \bibinfo{volume}{5}, \bibinfo{number}{1} (\bibinfo{year}{2012}),
  \bibinfo{pages}{3--14}.
\newblock


\bibitem[Dandekar(2023)]%
        {dandekar2023approach}
\bibfield{author}{\bibinfo{person}{Anaya Dandekar}.}
  \bibinfo{year}{2023}\natexlab{}.
\newblock \showarticletitle{An Approach for Anomaly Detection \& Prediction in
  Time-series Telecommunication Data}. In \bibinfo{booktitle}{\emph{2022 OPJU
  International Technology Conference on Emerging Technologies for Sustainable
  Development (OTCON)}}. IEEE, \bibinfo{pages}{1--6}.
\newblock


\bibitem[Doersch(2016)]%
        {doersch2016tutorial}
\bibfield{author}{\bibinfo{person}{Carl Doersch}.}
  \bibinfo{year}{2016}\natexlab{}.
\newblock \showarticletitle{Tutorial on variational autoencoders}.
\newblock \bibinfo{journal}{\emph{arXiv preprint arXiv:1606.05908}}
  (\bibinfo{year}{2016}).
\newblock


\bibitem[Hammerbacher et~al\mbox{.}(2021)]%
        {hammerbacher2021including}
\bibfield{author}{\bibinfo{person}{Tom Hammerbacher}, \bibinfo{person}{Markus
  Lange-Hegermann}, {and} \bibinfo{person}{Gorden Platz}.}
  \bibinfo{year}{2021}\natexlab{}.
\newblock \showarticletitle{Including Sparse Production Knowledge into
  Variational Autoencoders to Increase Anomaly Detection Reliability}. In
  \bibinfo{booktitle}{\emph{2021 IEEE 17th International Conference on
  Automation Science and Engineering (CASE)}}. IEEE,
  \bibinfo{pages}{1262--1267}.
\newblock


\bibitem[Higgins et~al\mbox{.}(2017)]%
        {higgins2017beta}
\bibfield{author}{\bibinfo{person}{Irina Higgins}, \bibinfo{person}{Loic
  Matthey}, \bibinfo{person}{Arka Pal}, \bibinfo{person}{Christopher Burgess},
  \bibinfo{person}{Xavier Glorot}, \bibinfo{person}{Matthew Botvinick},
  \bibinfo{person}{Shakir Mohamed}, {and} \bibinfo{person}{Alexander
  Lerchner}.} \bibinfo{year}{2017}\natexlab{}.
\newblock \showarticletitle{beta-vae: Learning basic visual concepts with a
  constrained variational framework}. In
  \bibinfo{booktitle}{\emph{International conference on learning
  representations}}.
\newblock


\bibitem[Intelligence(2023)]%
        {intelligence2023mobile}
\bibfield{author}{\bibinfo{person}{GSMA Intelligence}.}
  \bibinfo{year}{2023}\natexlab{}.
\newblock \showarticletitle{The mobile economy 2023}.
\newblock \bibinfo{journal}{\emph{London: GSM Association}}
  (\bibinfo{year}{2023}).
\newblock
\urldef\tempurl%
\url{https://www.gsma.com/mobileeconomy/wp-content/uploads/2023/03/270223-The-Mobile-Economy-2023.pdf}
\showURL{%
\tempurl}


\bibitem[Kingma and Welling(2013)]%
        {kingma2013auto}
\bibfield{author}{\bibinfo{person}{Diederik~P Kingma} {and}
  \bibinfo{person}{Max Welling}.} \bibinfo{year}{2013}\natexlab{}.
\newblock \showarticletitle{Auto-encoding variational bayes}.
\newblock \bibinfo{journal}{\emph{arXiv preprint arXiv:1312.6114}}
  (\bibinfo{year}{2013}).
\newblock


\bibitem[Li et~al\mbox{.}(2019)]%
        {li2019anomaly}
\bibfield{author}{\bibinfo{person}{Bing Li}, \bibinfo{person}{Shengjie Zhao},
  \bibinfo{person}{Rongqing Zhang}, \bibinfo{person}{Qingjiang Shi}, {and}
  \bibinfo{person}{Kai Yang}.} \bibinfo{year}{2019}\natexlab{}.
\newblock \showarticletitle{Anomaly detection for cellular networks using big
  data analytics}.
\newblock \bibinfo{journal}{\emph{IET Communications}} \bibinfo{volume}{13},
  \bibinfo{number}{20} (\bibinfo{year}{2019}), \bibinfo{pages}{3351--3359}.
\newblock


\bibitem[Locatello et~al\mbox{.}(2019)]%
        {locatello2019challenging}
\bibfield{author}{\bibinfo{person}{Francesco Locatello},
  \bibinfo{person}{Stefan Bauer}, \bibinfo{person}{Mario Lucic},
  \bibinfo{person}{Gunnar Raetsch}, \bibinfo{person}{Sylvain Gelly},
  \bibinfo{person}{Bernhard Sch{\"o}lkopf}, {and} \bibinfo{person}{Olivier
  Bachem}.} \bibinfo{year}{2019}\natexlab{}.
\newblock \showarticletitle{Challenging common assumptions in the unsupervised
  learning of disentangled representations}. In
  \bibinfo{booktitle}{\emph{international conference on machine learning}}.
  PMLR, \bibinfo{pages}{4114--4124}.
\newblock


\bibitem[MacQueen(1967)]%
        {macqueen1967classification}
\bibfield{author}{\bibinfo{person}{J MacQueen}.}
  \bibinfo{year}{1967}\natexlab{}.
\newblock \showarticletitle{Classification and analysis of multivariate
  observations}. In \bibinfo{booktitle}{\emph{5th Berkeley Symp. Math. Statist.
  Probability}}. University of California Los Angeles LA USA,
  \bibinfo{pages}{281--297}.
\newblock


\bibitem[Nediyanchath et~al\mbox{.}(2020)]%
        {9124843}
\bibfield{author}{\bibinfo{person}{Anish Nediyanchath}, \bibinfo{person}{Chirag
  Singh}, \bibinfo{person}{Harman~Jit Singh}, \bibinfo{person}{Himanshu
  Mangla}, \bibinfo{person}{Karan Mangla}, \bibinfo{person}{Manoj~K. Sakhala},
  \bibinfo{person}{Saravanan Balasubramanian}, \bibinfo{person}{Seema Pareek},
  {and} \bibinfo{person}{Shwetha}.} \bibinfo{year}{2020}\natexlab{}.
\newblock \showarticletitle{Anomaly Detection in Mobile Networks}. In
  \bibinfo{booktitle}{\emph{2020 IEEE Wireless Communications and Networking
  Conference Workshops (WCNCW)}}. \bibinfo{pages}{1--5}.
\newblock
\urldef\tempurl%
\url{https://doi.org/10.1109/WCNCW48565.2020.9124843}
\showDOI{\tempurl}


\bibitem[Niggemann and Lohweg(2015)]%
        {niggemann2015diagnosis}
\bibfield{author}{\bibinfo{person}{Oliver Niggemann} {and}
  \bibinfo{person}{Volker Lohweg}.} \bibinfo{year}{2015}\natexlab{}.
\newblock \showarticletitle{On the diagnosis of cyber-physical production
  systems}. In \bibinfo{booktitle}{\emph{Proceedings of the AAAI Conference on
  Artificial Intelligence}}, Vol.~\bibinfo{volume}{29}.
\newblock


\bibitem[Schiavone et~al\mbox{.}(2014)]%
        {schiavone2014diagnosing}
\bibfield{author}{\bibinfo{person}{Mirko Schiavone}, \bibinfo{person}{Peter
  Romirer-Maierhofer}, \bibinfo{person}{Pierdomenico Fiadino}, {and}
  \bibinfo{person}{Pedro Casas}.} \bibinfo{year}{2014}\natexlab{}.
\newblock \showarticletitle{Diagnosing device-specific anomalies in cellular
  networks}. In \bibinfo{booktitle}{\emph{Proceedings of the 2014 CoNEXT on
  Student Workshop}}. \bibinfo{pages}{18--20}.
\newblock


\bibitem[Shaikh et~al\mbox{.}(2022)]%
        {shaikh2022conceptual}
\bibfield{author}{\bibinfo{person}{Razin~A Shaikh},
  \bibinfo{person}{Sara~Sabrina Zemljic}, \bibinfo{person}{Sean Tull}, {and}
  \bibinfo{person}{Stephen Clark}.} \bibinfo{year}{2022}\natexlab{}.
\newblock \showarticletitle{The Conceptual VAE}.
\newblock \bibinfo{journal}{\emph{arXiv preprint arXiv:2203.11216}}
  (\bibinfo{year}{2022}).
\newblock


\bibitem[Sharma and Kapoor(2022)]%
        {sharma2022approach}
\bibfield{author}{\bibinfo{person}{Ashish Sharma} {and} \bibinfo{person}{Nitika
  Kapoor}.} \bibinfo{year}{2022}\natexlab{}.
\newblock \showarticletitle{Approach for Predicting Mobile Malware}. In
  \bibinfo{booktitle}{\emph{2022 4th International Conference on Advances in
  Computing, Communication Control and Networking (ICAC3N)}}. IEEE,
  \bibinfo{pages}{1614--1618}.
\newblock


\bibitem[Sohn et~al\mbox{.}(2015)]%
        {sohn2015learning}
\bibfield{author}{\bibinfo{person}{Kihyuk Sohn}, \bibinfo{person}{Honglak Lee},
  {and} \bibinfo{person}{Xinchen Yan}.} \bibinfo{year}{2015}\natexlab{}.
\newblock \showarticletitle{Learning structured output representation using
  deep conditional generative models}.
\newblock \bibinfo{journal}{\emph{Advances in neural information processing
  systems}}  \bibinfo{volume}{28} (\bibinfo{year}{2015}).
\newblock


\bibitem[Solyman and Yahya(2022)]%
        {solyman2022evolution}
\bibfield{author}{\bibinfo{person}{Ahmed Amin~Ahmed Solyman} {and}
  \bibinfo{person}{Khalid Yahya}.} \bibinfo{year}{2022}\natexlab{}.
\newblock \showarticletitle{Evolution of wireless communication networks: From
  1G to 6G and future perspective}.
\newblock \bibinfo{journal}{\emph{International Journal of Electrical and
  Computer Engineering}} \bibinfo{volume}{12}, \bibinfo{number}{4}
  (\bibinfo{year}{2022}), \bibinfo{pages}{3943}.
\newblock
\urldef\tempurl%
\url{https://doi.org/10.11591/ijece.v12i4.pp3943-3950}
\showURL{%
\tempurl}


\bibitem[Stoecklin(2006)]%
        {stoecklin2006anomaly}
\bibfield{author}{\bibinfo{person}{Marc Stoecklin}.}
  \bibinfo{year}{2006}\natexlab{}.
\newblock \showarticletitle{Anomaly detection by finding feature distribution
  outliers}. In \bibinfo{booktitle}{\emph{Proceedings of the 2006 ACM CoNEXT
  conference}}. \bibinfo{pages}{1--2}.
\newblock


\bibitem[Walker et~al\mbox{.}(2016)]%
        {walker2016uncertain}
\bibfield{author}{\bibinfo{person}{Jacob Walker}, \bibinfo{person}{Carl
  Doersch}, \bibinfo{person}{Abhinav Gupta}, {and} \bibinfo{person}{Martial
  Hebert}.} \bibinfo{year}{2016}\natexlab{}.
\newblock \showarticletitle{An uncertain future: Forecasting from static images
  using variational autoencoders}. In \bibinfo{booktitle}{\emph{Computer
  Vision--ECCV 2016: 14th European Conference, Amsterdam, The Netherlands,
  October 11--14, 2016, Proceedings, Part VII 14}}. Springer,
  \bibinfo{pages}{835--851}.
\newblock


\bibitem[Wang and Ferr{\'u}s(2021)]%
        {wang2021extracting}
\bibfield{author}{\bibinfo{person}{Shaoxuan Wang} {and} \bibinfo{person}{Ramon
  Ferr{\'u}s}.} \bibinfo{year}{2021}\natexlab{}.
\newblock \showarticletitle{Extracting cell patterns from high-dimensional
  radio network performance datasets using self-organizing maps and K-means
  clustering}.
\newblock \bibinfo{journal}{\emph{IEEE access}}  \bibinfo{volume}{9}
  (\bibinfo{year}{2021}), \bibinfo{pages}{42045--42058}.
\newblock


\end{thebibliography}

\newpage 
\clearpage
\appendix
\section{Appendix}

\begin{minipage}{\textwidth}
    \begin{figure}[H]
      \centering
      \includegraphics[width=0.575\linewidth]{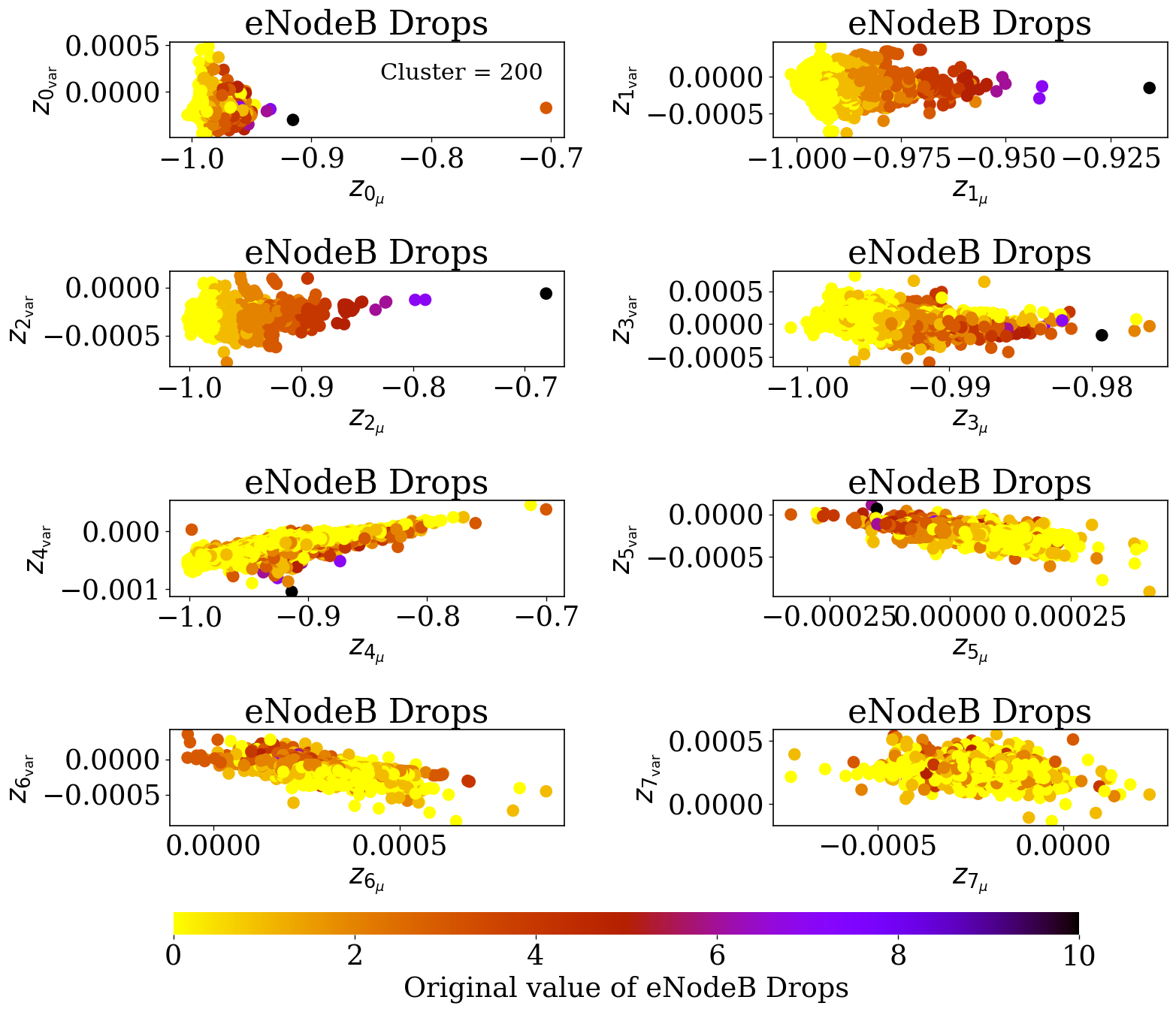}
      \caption{Latent representation of leaned means (x-axis) and log-variances (y-axis) predicted by the encoder for eNodeB Drops $(p(z2|c2))$ in training dataset for cluster 200}
      \label{fig:enod_fig}
    \end{figure}
    \begin{figure}[H]
      \centering
      \includegraphics[width=0.575\linewidth]{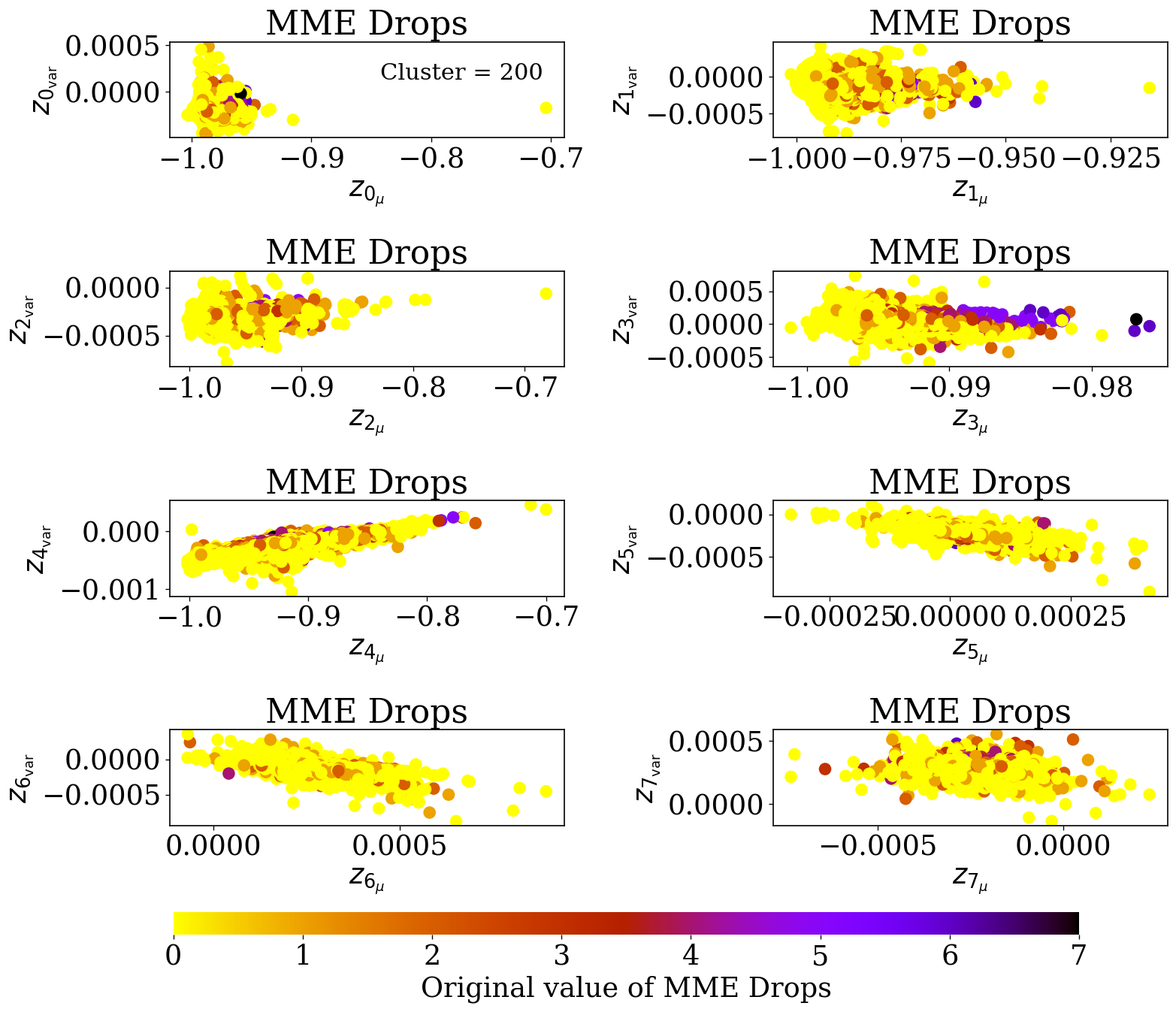}
      \caption{Latent representation of leaned means (x-axis) and log-variances (y-axis) predicted by the encoder for MME Drops $(p(z3|c3))$ in training dataset for cluster 200}
      \label{fig:mme_fig}
    \end{figure}
\end{minipage}

\end{document}